\documentclass[
]{ceurart}

\usepackage{wrapfig}
\usepackage{listings}
\usepackage{xcolor}
 \usepackage[normalem]{ulem}
\usepackage[utf8]{inputenc} 
\usepackage{tcolorbox}
\usepackage [english]{babel}
\usepackage [autostyle, english = american]{csquotes}
\usepackage{listings}
\usepackage{xcolor}
\MakeOuterQuote{"}

\usepackage{geometry}
\usepackage{caption}
\copyrightyear{2025}
\copyrightclause{Copyright for this paper by its authors.
  Use permitted under Creative Commons License Attribution 4.0
  International (CC BY 4.0).}

\conference{In: R. Campos, A. Jorge, A. Jatowt, S. Bhatia, M. Litvak (eds.): Proceedings of the Text2Story'25 Workshop, Lucca (Italy), 10-March-2025}

\title{Beyond Negation Detection: Comprehensive Assertion Detection Models for Clinical NLP}

\author[1]{Veysel Kocaman}[%
orcid=0000-0002-0065-6478,
email=veysel@johnsnowlabs.com]

\author[1]{Yigit Gul}[%
orcid=0009-0007-7184-9879,
email=yigit@johnsnowlabs.com]

\author[1]{M. Aytug Kaya}[%
orcid=0009-0009-7824-6389,
email=aytug@johnsnowlabs.com]

\author[1]{Hasham Ul Haq}[%
orcid=0000-0002-8417-3288,
email=hasham@johnsnowlabs.com]

\author[1]{Mehmet Butgul}[%
email=mehmet@johnsnowlabs.com]

\author[1]{Cabir Celik}[%
email=cabir@johnsnowlabs.com]

\author[1]{David Talby}[%
orcid=0000-0003-2782-5478,
email=david@johnsnowlabs.com]

\address[1]{John Snow Labs inc. 16192 Coastal Highway,
Lewes, DE 19958, USA}

\begin{document}
\maketitle
\begin{abstract}
Assertion status detection is a critical yet often overlooked component of clinical NLP, essential for accurately attributing extracted medical facts. Past studies narrowly focused on negation detection, resulting in underperforming commercial solutions such as AWS Medical Comprehend, Azure AI Text Analytics, and GPT-4o due to their limited domain adaptation. To address this gap, we developed state-of-the-art assertion detection models, including fine-tuned LLMs, transformer-based classifiers, few-shot classifiers, and deep learning (DL) approaches and evaluated our models against cloud-based commercial API solutions and legacy rule-based NegEx approach as well as GPT-4o. Our fine-tuned LLM achieves the highest overall accuracy (0.962), outperforming GPT-4o (0.901) and commercial APIs by a notable margin, particularly excelling in \textit{Present} (+4.2\%), \textit{Absent} (+8.4\%), and \textit{Hypothetical} (+23.4\%) assertions. Our DL-based models surpass commercial solutions in \textit{Conditional} (+5.3\%) and \textit{Associated with Someone Else} (+10.1\%), while few-shot classifier offers a lightweight yet highly competitive alternative (0.929), making it ideal for resource-constrained environments. 
Integrated within Spark NLP, our models consistently outperform black-box commercial solutions while enabling scalable inference and seamless integration with medical NER, Relation Extraction, and Terminology Resolution. These results reinforce the importance of domain-adapted, transparent, and customizable clinical NLP solutions over general-purpose LLMs and proprietary APIs.

\end{abstract}







\section{Introduction}
The widespread adoption of Electronic Health Records (EHRs) has transformed healthcare, with 96\% of non-federal acute care hospitals and 78\% of office-based physicians in the United States using certified EHR systems by 2021. This digitization has created vast patient data repositories, opening new avenues for clinical applications and research. \cite{ONC2022}.
To harness this valuable information and discover patterns in EHRs, various Natural Language Processing (NLP) tasks have been performed. Among these, the classification of assertions stands out as a critical but understudied task. Accurate assertion classification allows for the determination of whether a medical concept is \textit{present},\textit{absent}, \textit{possible}, \textit{hypothetical}, \textit{conditional}, or \textit{associated with someone} other than the patient, crucial for extracting actionable insights from EHRs, driving clinical decision-making, and facilitating healthcare analytics \cite{wen2024case}. In other words, the status of an assertion explains how
a named entity (e.g. clinical finding, procedure, lab result)
pertains to the patient by assigning a label such as \textit{present}
(”patient is diabetic”), \textit{absent} (”patient denies nausea”), \textit{conditional} (”dyspnea while climbing stairs”), or \textit{associated with
someone else} (”family history of depression”). Table \ref{tab:assertion_categories} illustrates different assertion classes with their label distribution and sizes.

Although early studies often equated assertion detection with negation detection in time, sophisticated machine learning and deep learning methodologies evolved rudimentary rule-based approaches. Early techniques such as NegEx \cite{chapman2001simple}, ConText \cite{chapman2007context}, NegFinder \cite{mutalik2001use} and NegExpander \cite{aronow1999ad} relied on hand-crafted rules and regular expressions, achieving high precision but suffering from low recall due to rigid patterns \cite{Perez2023}. In order to learn more about rule-based approaches towards assertion detection, a reader is advised to check out a comprehensive study \cite{uzuner2009machine} evaluating these approaches in detail.  
Deep learning methods, particularly transformer-based models and attention mechanisms, emerged as powerful alternatives, offering more nuanced understanding of clinical text. However, these approaches consistently faced challenges such as requiring large annotated datasets and struggling with minority classes, especially in detecting \textit{possible} medical assertions. Recent developments have focused on addressing these limitations through innovative approaches like multi-task learning, pre-training techniques, and Large Language Models (LLMs). 

Bhatia et al. demonstrated the effectiveness of a multitask learning approach for jointly modeling named entity recognition and negation assertion in clinical texts. By utilizing shared parameters, their model achieved improved contextual representation and overcame challenges associated with neural networks in negation detection, outperforming rule-based systems in conjunction with the proposed conditional softmax decoder \cite{Bhatia2019}.

Chen et al. explored applying attention-based bi-LSTM architectures for negation and assertion detection in clinical notes, leveraging the ability to selectively focus on relevant information and automatically capture semantic details without relying on external knowledge inputs \cite{Chen2019}.

Aken et al. proposed a comprehensive study of clinical assertion detection models by manually annotating 5,000 assertions in the MIMIC-III dataset, evaluating medical language models' performance and transferability across different medical domains, and releasing their annotated dataset to address label sparsity and diversity challenges in existing research \cite{van-aken-etal-2021-assertion}. Similarly, Wang et al. proposed a novel prompt-based learning approach for assertion classification that addresses existing limitations by leveraging few-shot learning and advanced reasoning techniques \cite{WANG2022104139}.

Yuan et al. proposed a deep learning approach for automatic Electronic Medical Record (EMR) sectioning using MIMIC-III data, developing hand-crafted rules to create gold-standard labels and generating multiple note versions with varied section heading formats to train models that achieve robust adaptability and high accuracy in EMR segmentation \cite{yuan2024blackboxsegmentationelectronicmedical}.

Ji et al. proposed a novel method leveraging Large Language Models (LLMs) with advanced reasoning techniques like Tree of Thought (ToT), Chain of Thought (CoT), and Self-Consistency (SC), combined with Low-Rank Adaptation (LoRA) fine-tuning, to transform assertion detection into a generative task that enables more nuanced, contextually aware, and data-efficient medical text understanding across multiple assertion categories \cite{ji2024assertiondetectionlargelanguage}. 


While existing clinical NLP approaches have made significant strides in assertion detection, predominantly focusing on negation, they have consistently fallen short of providing a comprehensive, multi-category framework capable of robustly addressing the full spectrum of medical concept assertions. In this paper, we present a comprehensive implementation of assertion detection within Healthcare NLP library \cite{kocaman2022accurate} (based on Spark NLP \cite{kocaman2021spark} \cite{haq2021deeperclinicaldocumentunderstanding} \cite{haq2022miningadversedrugreactions} ecosystem), utilizing state-of-the-art models and annotators to achieve high accuracy and efficiency in clinical NLP tasks. Our approach transcends traditional negation detection methods by offering a comprehensive, fully integrable end-to-end solution that addresses the entire spectrum of assertion types, including \textit{present, absent, possible, hypothetical, conditional, and assertions associated with someone other than the patient}. This holistic method leverages advanced deep learning architectures, few-shot learning techniques, and flexible rule-based systems to overcome common challenges in clinical texts, such as class imbalance and ambiguous concept expressions. Specifically, we explore the following architectures/ modules that we developed during this study to detect assertion status from clinical notes:

\begin{table}[ht]
\caption{Examples from the i2b2 dataset illustrating different assertion classes with their label distribution.}
\centering
\resizebox{\textwidth}{!}{%
\label{tab:assertion_categories}
\begin{tabular}{|p{10cm}|p{2cm}|p{4cm}|l|}
\hline
\textbf{Text} & \textbf{Label} & \textbf{Description} & \textbf{Size} \\
\hline
Overnight, the patient became \colorbox{yellow}{\textbf{hypoxic}}, dropping to the 80 's. & \textit{present} & Confirms the presence of a medical condition. & 8622 \\ \hline
He gets  \colorbox{yellow}{\textbf{short of breath}} with one flight of stairs. & \textit{conditional} & Represents conditions that might occur under specific circumstances or conditions. & 148 \\ \hline
Small stroke, nearly recovered, likely \colorbox{yellow}{\textbf{embolic from carotid artery}}. & \textit{possible} & Suggests uncertainty or potential presence of a condition. & 652 \\ \hline
There was no evidence of \colorbox{yellow}{\textbf{diarrhea}} during medical Lawrence Memorial Hospital stay. & \textit{absent} & Indicates the negation or nonexistence of a medical condition. & 2594 \\ \hline
Mother suffer \colorbox{yellow}{\textbf{MI}} in her 50 's, died at age 59. & \textit{associated with someone else (awse)} & Refers to medical conditions related to individuals other than the patient, such as family members. & 131 \\ \hline
Hydrocodone 5 mg with Tylenol , one to two tablets every four hours p.r.n. \colorbox{yellow}{\textbf{pain}}. & \textit{hypothetical} & Denotes speculative or conjectural conditions that are not currently present. & 445 \\ \hline
\end{tabular}%
}
\end{table}




\begin{itemize}
    \item \textbf{Assertion Detection with LLMs}: 
    To overcome the limitations coming from data collection and annotations to design ML/ DL based assertion detection models, we experiment with leveraging LLMs pretrained on extensive medical datasets to enhance assertion detection accuracy and comprehensiveness in zero shot settings.
    \item \textbf{Assertion Detection with a DL Model}: A deep learning-based annotator built on a Bi-LSTM architecture, inspired by \cite{fancellu-etal-2016-neural}. This model processes medical concepts and their surrounding tokens using word embeddings within a defined scope window. 
    \item \textbf{Assertion Detection with a Bert For Sequence Classification (BFSC)}: This approach leverages a transformer-based model, BERT, to classify assertion status in medical texts. By encoding the contextual relationships within sequences, BERT enables accurate detection of negations, affirmations, and other assertion types. 
    \item \textbf{Few Shot Assertion Detection with Transformers}: A few-shot learning-based classifier that combines sentence embeddings with lightweight classification models to achieve high accuracy with minimal training data.  
    \item \textbf{Rule-based Assertion Detection with Contextual Awareness}: A rule-based annotator designed to enhance assertion detection accuracy in complex clinical contexts. By leveraging customizable keyword sets, regex patterns, and scope windows, this model adapts to diverse clinical scenarios. 
\end{itemize}

The subsequent sections will systematically evaluate these assertion detection architectures on a well-known benchmark dataset and compare them with GPT4o, a rule-based algorithm (NegEx), and cloud-based healthcare-specific APIs offered by commercial providers (AWS Medical Comprehend and Azure AI Text Analytics for Health). Our analysis will showcase a novel combined pipeline that integrates these models, demonstrating their complementary strengths in enhancing assertion detection performance across various computational paradigms and clinical scenarios.

\section{Methodology}

In this section, we explain the details of various model architectures supported and shipped as a pretrained model into Healthcare NLP library by John Snow Labs (JSL).




\subsection{Assertion Detection with LLMs}
Traditional approaches to assertion detection in medical text, such as rule-based NLP systems and machine learning or deep learning models, often require significant manual effort to design patterns and frequently fail to capture less common assertion types, resulting in incomplete contextual understanding. To overcome these limitations, we explored finetuning an LLM with assertion detection datasets to enhance assertion detection accuracy and comprehensiveness.

 We explored training LLama-3.1-8B\cite{grattafiori2024llama3herdmodels} on the i2b2 assertion dataset. We fine tuned LLama-3.1-8B\cite{grattafiori2024llama3herdmodels} model with the i2b2 assertion training dataset using LoRA fine-tuning \cite{wang2023lora} approach without quantization. LoRA offers parameter efficiency by updating only a small subset of parameters, reducing memory and computational overhead. It minimizes overfitting risk by keeping pre-trained weights fixed, which makes it ideal for small training datasets and it preserves pre-trained knowledge, maintaining generalization capabilities while allowing task-specific tuning. Our final configuration used a LoRA rank of 16, LoRA alpha of 32, and 5 training epochs.

For fine-tuning, a simple and efficient prompt structure is of paramount importance. We explicitly included a detailed description of each assertion status to evidently improve performance. Additionally, we replaced the term \textit{Present} with \textit{Confirmed} which yielded better results, likely due to improved clarity and alignment with the task's semantics. Including descriptions of assertion statuses in the input prompt also allowed for minor adjustments during inference, enhancing flexibility and adaptability.
Our experimentation proved counterintuitive when we studied the context: inputting the whole document created complexity and confusion, impairing performance. We replaced this approach with a context windowing strategy, extracting two sentences before and after the target text. This strategy substantially reduced training time and increased the model's ability to focus on relevant information.

\subsection{Assertion Detection via DL Model}

Assertion Detection via DL Model (AssertionDL) is a classification model based on a Bi-LSTM framework, representing a modified version of the architecture proposed by \cite{fancellu-etal-2016-neural}. In this implementation, entities (also referred to as chunks) are processed alongside a context string. The context string and entities are tokenized and embedded before being passed to the Bi-LSTM model. It is important to balance the length of the context string, as excessively long sequences can result in vanishing gradients, which may hinder the model’s performance.

An analysis of the i2b2 dataset revealed that 95\% of the relevant scope tokens (neighboring words) are located within a window spanning 9 tokens to the left and 15 tokens to the right of the target tokens. Based on this observation, we adopted the same window size for our model. 

The model has been implemented in Healthcare NLP library as an annotator called \textit{AssertionDLModel}, enabling seamless integration into the Spark NLP library for clinical and biomedical text processing.


\subsection{Assertion Detection via Bert For Sequence Classification (BFSC)}




While the LLM approach generates new tokens as part of its output, we also explored a more direct approach by framing the problem as a classification task. In this setup, the input consists of the entity chunk and its surrounding context, while the output is the predicted assertion status class. Specifically, we implemented a classification layer on top of a transformer model, such as BERT\cite{devlin2019bert}, to perform assertion status prediction, a technique known as BERT for Sequence Classification.

Rather than using the standard BERT model, we utilized the pre-trained BERT models from \cite{alsentzer2019publiclyavailableclinicalbert}, which have been fine-tuned on biomedical text. Among these, we selected the model trained on BioBert \cite{lee2020biobert}, as it demonstrated the best performance for our task. This approach has previously shown promising results for assertion detection \cite{van-aken-etal-2021-assertion} and helps the model focus on the target entity, even in contexts that contain multiple entities.

The input text was prepared by a novel approach as explained in \cite{van-aken-etal-2021-assertion}.
In addition, we experimented with varying context lengths by incorporating additional sentences around the target chunk. However, this approach yielded minimal performance improvements while evidently increasing training and processing time.


\subsection{Few Shot Assertion Detection via Transformers}

Few Shot Assertion Detection via Transformers (FewShotAssertion) in this study is built on a modified version of SetFit (Sentence Transformer Fine-Tuning) framework \cite{tunstall2022efficientfewshotlearningprompts}, which leverages sentence-transformer embeddings and a lightweight classifier for few-shot learning. SetFit enables efficient fine-tuning by coupling a pre-trained sentence-transformer model with a classifier trained on task-specific data using contrastive learning. 

The model takes as input the assertion context and the target entity, embedding them using a pre-trained transformer encoder. These embeddings are then fine-tuned using contrastive learning to align positive examples while separating negative ones in the embedding space. A lightweight linear classifier is subsequently trained on the refined embeddings to predict the assertion status. This approach is particularly well-suited for assertion detection in the i2b2 dataset, as it effectively handles limited labeled data while maintaining robust performance.



\subsection{Rule-based Assertion Detection with Contextual Awareness (ContextualAssertion)}

The model based on this architecture enables assertion detection by labeling entities (chunks) based on user-defined rules and contextual patterns, building upon principles similar to ConText \cite{harkema2009context} and the widely used NegEx framework \cite{Chapman2001ASA}. Unlike NegEx, which focuses on negation detection using fixed lexical patterns, the Contextual Assertion module provides advanced configurability through prefix and suffix keywords, regex patterns, exception handling, and customizable scope windows. These enhancements enable the establishment of complex linguistic rules, allowing the annotator to function as a robust and flexible guardrail for NLP pipelines. Following are some of its features:





\label{sec:pipeline}
\subsection{Using Assertion Detection Models within Healthcare NLP Pipeline}

While the i2b2 dataset provides pre-annotated named entities (including their indices), practical applications require extracting these entities directly from unstructured text. To address this, we propose an end-to-end, flexible pipeline with component sharing, as illustrated in Figure \ref{fig:diagram}.

In this pipeline, named entities are identified using Healthcare NLP's NER models and subsequently passed to assertion models for assertion status detection. The assertion model utilizes the same embeddings as the NER model, enabling embedding sharing for improved memory management and reduced latency.

The pipeline also supports a stacking approach, allowing multiple assertion models to coexist within a single framework. To enhance performance, we developed a merging mechanism that combines predictions from three assertion models and prioritizes them to produce a unified label for each entity based on the performance of each assertion models on certain entities. The key components of this pipeline includes \textbf{AssertionDL}, \textbf{FewShotAssertion} and \textbf{ContextualAssertion}.


    

To resolve conflicts in predictions across models, a \textbf{majority voting mechanism} is applied. This approach ensures the final label reflects the consensus among models, mitigating the impact of outlier predictions.

\subsection{Pretrained Models Offered in Healthcare NLP}


The Healthcare NLP library by JSL offers a range of domain-specific pretrained assertion models (e.g., oncology, radiology) that have been fine-tuned or trained using the architectures explored in this study. These models are fully optimized for integration within a Healthcare NLP pipeline, enabling scalable and efficient deployment. For a detailed list of pretrained clinical assertion models and their corresponding benchmarks, refer to the Table \ref{tab:pretrained_models_comparison_transposed} that showcases the best performance scores achieved by these models across multiple assertion categories (12 categories, more than what is covered in this study) including \textit{Present}, \textit{Past}, \textit{Possible}, \textit{Absent}, \textit{Hypothetical}, \textit{Family}, \textit{Someone Else}, \textit{Planned}, \textit{Conditional}, \textit{Confirmed}, \textit{Negative}, and \textit{Suspected}.






\section{Experiments and Results}

\subsection{Experimental Setup}



In this study, we benchmarked the performance of our assertion classification approaches — \texttt{AssertionDL}, \texttt{FewShotAssertion}, \texttt{ContextualAssertion}, \texttt{BFSC} and a combined pipeline against available counterparts -- NegEx, AWS Comprehend Medical, Azure AI Text Analytics, and GPT-4o.

NegEx is a rule-based algorithm designed to identify negation in clinical text, particularly to determine whether a medical concept is \textit{absent} or not. Introduced by \cite{chapman2001simple}, NegEx uses regular expressions and predefined linguistic patterns to detect negation cues (e.g., ``no,'' ``denies'') and their scope within a sentence. 

GPT-4o was employed to benchmark assertion detection for medical conditions in the i2b2 dataset. As the disclosure statement of i2b2 dataset prohibits sharing the data via cloud based APIs, we obfuscated the i2b2 dataset both for PHI and medical terms using Healthcare NLP tools provided by John Snow Labs, and then run the evaluation. A carefully crafted prompt (see figure \ref{fig:gpt_prompt}) guided the model to classify assertion statuses for specified medical entities.

\textbf{AWS Comprehend Medical} is an NLP service offered by Amazon Web Services, designed to automate the extraction of medical information from unstructured text. 
\textbf{Azure AI Text Analytics} is a natural language processing (NLP) service provided by Microsoft, designed to analyze and extract insights from unstructured text. 

Both AWS and Azure services extract entities at first and then annotate them with assertion labels (e.g., \textit{present}, \textit{absent}, \textit{hypothetical}). We aligned these annotations with i2b2 dataset taxonomies via label mapping to ensure consistency in evaluation. Since these services assign assertion labels only to the entities extracted by them at first, the evaluation is run over the partially or fully overlapped common entities from i2b2 dataset. The overlapping rates can be seen at Table \ref{Tab:mathing_rates} in Appendix. 

To maintain consistency, labels from Azure AI and AWS Comprehend were mapped to i2b2 equivalents. Matches were categorized into Full Match, Partial Match, and No Match, focusing the evaluation on full and partial matches. Statistics for matching outcomes are summarized in Table \ref{Tab:mathing_rates} with label mapping details available in Table \ref{tab:comprehensive_label_mapping} in Appendix for a reference.

\subsection{Dataset Description}
The evaluation and benchmarking in this study are conducted exclusively on the \textbf{official 2010 i2b2 dataset (test split)}\cite{i2b2}, which represents a comprehensive resource for assessing assertion detection frameworks in real-world clinical scenarios. The results focus on both individual models and combined pipelines, showcasing their relative strengths and collective impact on performance.



The dataset utilized in this study \textbf{covers all six assertion categories:} \textit{Absent}, \textit{Associated with someone else}, \textit{Conditional}, \textit{Hypothetical}, \textit{Possible}, and \textit{Present}. However, the fine-tuned LLM excludes the \textit{Conditional} label due to its ambiguity with the \textit{Hypothetical} label, which could complicate fine-tuning. This exclusion simplifies training and sharpens the model's focus on the remaining categories. In contrast, other models, including LLMs, retain all six categories to ensure a thorough evaluation of performance across the full range of assertion types.

\subsection{Comparative Results}

Table \ref{tab:overall_benchmarks} presents the experimental results, highlighting the performance of each model across relevant categories.

\begin{table}[hb!]
\scriptsize
\centering
\caption{Comparison of assertion models across various categories. Best performing model for each category is represented with bold characters.  The models in the first section of this table are developed by JSL. In LLM and GPT-4o experiments, \textit{hypothetical} and \textit{conditional} labels are merged/treated as a single label. 
}

\label{tab:overall_benchmarks}

\resizebox{\textwidth}{!}{
\begin{tabular}{l|c|c|c|c|c|r!{\vrule width 1.5pt}c}

\hline
\textbf{Model}             & \textbf{present} & \textbf{absent} & \textbf{possible} & \textbf{hypothetical} & \textbf{conditional} & \textbf{awse*} & \textbf{weighted avg} \\
\hline

\textit{\textbf{Combined Pipeline**}} & 0.963             & 0.951            & 0.755              & 0.875                  & 0.511                & 0.922       &0.941   \\
\textit{\textbf{AssertionDL}}       & 0.941             & 0.898            & 0.672              & 0.761                  & \textbf{0.599}        & 0.886         &0.907\\
\textit{\textbf{FewShotAssertion}}  & 0.955             & 0.942            & 0.748              & 0.872                  & 0.293                & 0.809         &0.929 \\
\textit{\textbf{ContextualAssertion}} & -              & 0.929            & 0.708              & -                     & -                & 0.835               &0.883 \\
\textbf{Fine Tuned LLM}    & \textbf{0.976}    & \textbf{0.975}   & 0.759     & 0.911          & -                  & \textbf{0.943}                            &\textbf{0.962}  \\
\textbf{BFSC (BioBert)} & 0.975 & 0.972 & \textbf{0.787} & 0.918 & 0.590 & 0.913        &0.957 \\
\noalign{\hrule height 1pt}
\textbf{GPT-4o}            & 0.937             & 0.891            & 0.692              & 0.677                  & -                   & 0.805           &0.901  \\
\textbf{Azure Ai Text Analytics} & -           & 0.761            & 0.583              & 0.763                  & 0.569                & 0.800          &0.727  \\
\textbf{AWS Med Comprehend}    & 0.882             & 0.788            & 0.659              & 0.617                  & -                   & 0.737       &0.839 \\
\textbf{NegEx}             & -                & 0.897            & -                 & -                     & -                   & -                  &0.897 \\
\noalign{\hrule height 1pt}
\textbf{BFSC latest best \cite{van-aken-etal-2021-assertion}}    & 0.979            & 0.972            & 0.786              & -                  & -                   & -   &0.952\\
\textbf{Prompt-based Bert \cite{wang2022trustworthy}}    & 0.971             & 0.968            & 0.763              & \textbf{0.921}                  & 0.485                  & 0.875  &0.951 \\

\hline
\end{tabular}
}
\begin{flushleft}
\footnotesize{\textit{*awse: associated with someone else.}}
\footnotesize{\textit{**Combined pipeline elements denoted in \textit{italics}.}}
\footnotesize{\textit{***BFSC latest best is benchmarked only on 3 labels by its authors \cite{van-aken-etal-2021-assertion}; hence excluded from comparison.}}

\end{flushleft}
\end{table}

Our fine-tuned LLM, based on the LLaMA 3.1-8B model and trained using LoRA on the i2b2 dataset, demonstrates superior performance in most categories compared to other models. This approach aligns with recent research in domain adaptation for clinical NLP tasks \cite{zhao2024loraland}. The results emphasize the efficacy of smaller, domain-specific models, which, when coupled with carefully engineered prompts, can often outperform much larger, general-purpose models. Our experimental findings indicate near-perfect performance across most categories, with only minor underperformance in the \textit{possible} and \textit{hypothetical} labels. Notably, our model excels not only in covering a broader range of categories but also in evidently outperforming commercial solutions such as GPT-4o, Azure AI Text Analytics, and AWS Comprehend.
    
The combined pipeline, which integrates rule-based methods with machine learning techniques, closely mirrors the performance of the fine-tuned LLM across most categories. This hybrid approach, which captures the strengths of both deep learning and rule-based systems, outperforms comparable solutions offered by Azure and AWS in every category except the \textit{conditional} label. Unlike Azure AI Text Analytics for Health and AWS Medical Comprehend, which are API-based black-box solutions, our pipeline offers customization and fine-tuning options, allowing for potential performance improvements across all categories, including the \textit{conditional} label. This flexibility represents a significant advantage in adapting the system to meet specific healthcare needs and optimizing performance across various clinical NLP tasks.

For use cases where deploying the full combined pipeline is not feasible, users can still achieve exceptional results by leveraging its individual components. AssertionDL, in particular, stands out as a versatile solution, effectively handling all assertion categories with its advanced deep learning architecture. It performs particularly well in the \textit{conditional} and \textit{associated with someone else} categories, demonstrating superior results in the \textit{conditional} label. Notably, AssertionDL outperforms GPT-4o in most categories, making it a robust standalone option for clinical assertion tasks.
    
The FewShotAssertion model can be used both standalone and as part of the pipeline, offering an ideal solution for rapid training and inference in resource-constrained clinical NLP environments where efficiency is crucial. It performs comparably to the fine-tuned LLM across most categories, with the exception of the "conditional" category. However, when integrated into the Healthcare NLP pipeline, its contribution of \textit{absent} and \textit{hypothetical} labels helps mitigate this limitation.

The BFSC model highlights the power of domain adaptation in clinical NLP tasks. By leveraging the domain-specific BioBERT language model and employing a sequence classifier, this approach demonstrates superior performance due to its fine-tuning on meticulously curated training data. While the BFSC model slightly underperforms compared to AssertionDL in the \textit{conditional} label category, its performance is close to the benchmark, and it holds potential for further improvement through strategic augmentation of the training dataset.

Despite its superior performance, LLM-based solutions come with substantial computational costs, requiring GPUs to run efficiently while still being slower. In our benchmarks, what takes around 3 seconds using our deep-learning-based approach on a CPU requires around 300 seconds on a GPU-powered LLM, 
which is 100× slower. \textbf{Given that GPU instances cost more than CPU instances—often 10–50× higher per hour—the operational cost of running LLM-based assertion detection can be thousands of times more expensive for only a 1-2\% accuracy gain.} 
This highlights the trade-off between accuracy and feasibility, where our lightweight, domain-adapted models provide a far more scalable and cost-effective alternative for real-world clinical NLP applications (see Table \ref{tab:speed_benchmarks}).
    



\section{Limitations}

While this study demonstrates notable advancements in clinical assertion detection, several limitations should be acknowledged. The models were benchmarked exclusively on the i2b2 dataset, which may limit generalizability to diverse clinical contexts. However, beyond this study, there are numerous models that have been trained for use in various domains such as oncology and radiology. Their F1 scores are available in Section A.5, Pre-trained Assertion Models in Healthcare NLP, and Table A4. These models can be accessed via the JohnSnowLabs Model Hub page\cite{johnsnowlabsModelHub}. Performance on underrepresented assertion types (e.g., conditional, associated with someone else) could vary in real-world settings with different label distributions. 

Although the fine-tuned LLM achieved state-of-the-art accuracy, its GPU dependency and 100× slower inference speed compared to CPU-based DL models raise practical scalability concerns, potentially hindering deployment in resource-constrained healthcare environments. Despite addressing label skew (e.g., absent and present dominate the dataset), minority classes like hypothetical (3.5\% prevalence) and conditional (1.2\%) still showed lower F1 scores, suggesting residual bias in model predictions.

Commercial APIs (AWS, Azure) were evaluated only on overlapping entities detected by their proprietary NER systems, introducing selection bias, and partial matches (28–35\% of cases) may have skewed performance metrics for these systems. 
 
\section{Ethical Considerations}

The development of clinical assertion detection models necessitates ethical scrutiny due to potential biases, privacy risks, and implications for patient care. The i2b2 dataset may contain demographic biases, risking inequitable model performance across populations. Future work should incorporate fairness audits and demographic stratification to mitigate these risks. Privacy concerns arise from processing sensitive patient data, particularly with cloud-based APIs (e.g., GPT-4o, AWS, Azure), necessitating transparent data governance frameworks for compliance with HIPAA and GDPR. The lack of interpretability in black-box models threatens clinical trust, underscoring the need for explainability tools to audit model decisions. Over-reliance on automation may lead to uncritical adoption in healthcare workflows, necessitating human-in-the-loop validation mechanisms. Additionally, the high computational cost of LLM training raises sustainability concerns, warranting efficiency-focused approaches. Future research should prioritize bias mitigation, open fairness benchmarks, ethical model documentation, and federated learning to enhance privacy and equity.

\section{Conclusion}\label{conclusion}



In this study, we present a comprehensive evaluation of JSL’s state-of-the-art assertion detection models, covering architectures from lightweight deep learning (DL) models to advanced fine-tuned LLMs. Overall, our fine-tuned LLM achieves the highest overall accuracy (0.962), outperforming GPT-4o (0.901) and commercial APIs by a notable margin, particularly in \textit{Present}, \textit{Absent}, and \textit{Hypothetical} assertions. However, this comes at a high computational cost; 
Our DL-based models run 100× faster on a CPU than the LLM on a GPU, while the LLM is thousands of times more expensive for just 1-2\% better accuracy. This highlights the impracticality of LLM-based assertion detection for real-time, scalable clinical NLP.

Our AssertionDL and FewShotAssertion models provide strong, efficient alternatives, excelling in categories like \textit{Conditional} and \textit{Associated with someone else} assertions, while BFSC achieves near-parity with our fine-tuned LLM. The Combined Pipeline outperforms all commercial solutions and offers a balance of accuracy and efficiency. As part of a scalable, production-ready Healthcare NLP library, these models seamlessly integrate with other clinical NLP components, enabling robust, high-performance assertion detection at scale. \textbf{Our results highlight that smaller, domain-specific models outperform commercial black-box solutions like GPT-4o, Azure AI, and AWS Medical Comprehend in both accuracy and scalability.}
Integrated within Spark NLP, our pretrained assertion models and model architectures provide production-ready, cost-effective alternatives for clinical text analysis, filling a critical gap in extracting accurate medical insights.



\break
\bibliography{references}
\newpage

\appendix
\renewcommand{\thetable}{A\arabic{table}} 
\renewcommand{\thefigure}{A\arabic{figure}}
\setcounter{table}{0} 
\setcounter{figure}{0} 

\section*{Appendix}

\subsection*{A.1 A Spark NLP pipeline}

\begin{figure}[htbp]
    \centering
    \caption{The flow diagram of a Spark NLP pipeline. When we fit() on the pipeline with a Spark data frame, its text column is fed into the DocumentAssembler() transformer and a new column document is created as an initial entry point to Spark NLP for any Spark data frame. Then, its document column is fed into the SentenceDetector(), Tokenizer() and WordEmbeddings(). Now data is ready to be fed into NER models and then to the assertion model.
    }
    \includegraphics[width=1\textwidth]{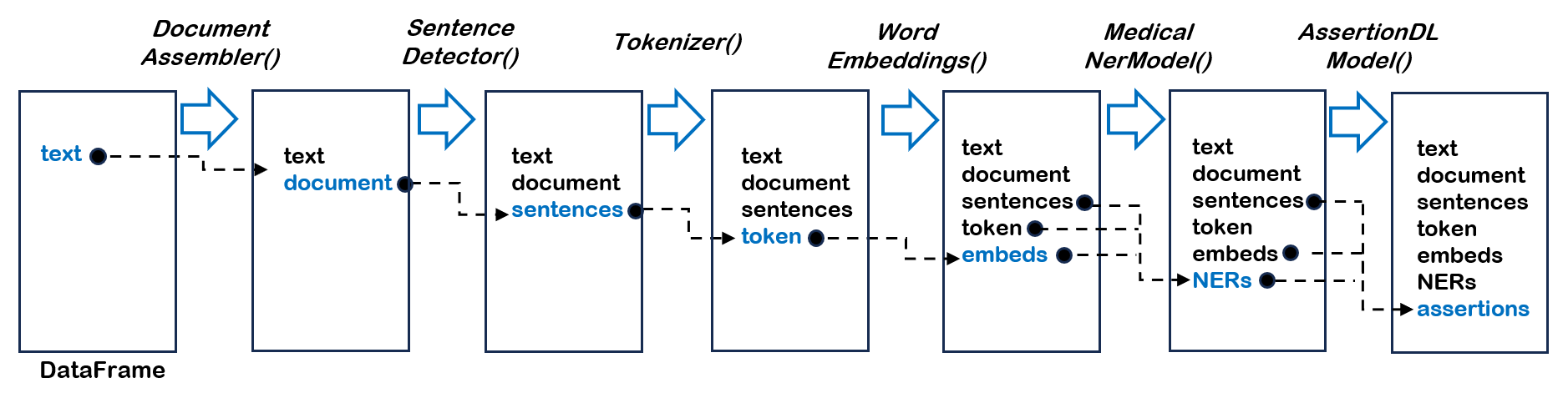}    
    \label{fig:diagram}
\end{figure}

\subsection*{A.2  Entity Overlapping Rates}
\begin{table}[htb!]
\label{tab:A1 entity overlapping}
\normalsize
\centering
\caption{Entity overlapping rates across the methods evaluated in this study. Percentages are calculated relative to Total Rows.}
\label{tab:assertion_model_comparison}
\resizebox{\textwidth}{!}{
\begin{tabular}{|l|p{5cm}|p{1.7cm}|p{2cm}|p{2cm}|p{2cm}|p{2cm}|p{2cm}|p{1cm}}
\hline
\textbf{Model} & \textbf{Predicted Labels} & \textbf{Full Match (\%)} & \textbf{Partial Match (\%)} & \textbf{No\ \ \ \ \ \ Match (\%)} & \textbf{Predicted Rows} & \textbf{Total Rows} \\
\hline
Fine-Tuned LLM & present, absent, possible, hypothetical, associated with someone else & 12444 (100.00\%) & 0\ \ \ \ \ \ \ \ \ \ \ \  (0.00\%) & 0\ \ \ \ \ \ \ \ \ \ \ \  (0.00\%) & 12444 & 12444 \\\hline
BFSC (BioBert) & present, absent, possible, hypothetical, associated with someone else & 12592 (100.00\%) & 0\ \ \ \ \ \ \ \ \ \ \ \  (0.00\%) & 0\ \ \ \ \ \ \ \ \ \ \ \  (0.00\%) & 12592 & 12592 \\\hline
AssertionDL & present, absent, possible, hypothetical, conditional, associated with someone else & 12592 (100.00\%) & 0\ \ \ \ \ \ \ \ \ \ \ \  (0.00\%) & 0\ \ \ \ \ \ \ \ \ \ \ \  (0.00\%) & 12592 & 12592 \\\hline
FewShotAssertion & present, absent, possible, hypothetical, conditional, associated with someone else & 12592 (100.00\%) & 0\ \ \ \ \ \ \ \ \ \ \ \  (0.00\%) & 0\ \ \ \ \ \ \ \ \ \ \ \  (0.00\%) & 12592 & 12592 \\\hline
Combined Pipeline & present, absent, possible, hypothetical, conditional, associated with someone else & 12592 (100.00\%) & 0\ \ \ \ \ \ \ \ \ \ \ \  (0.00\%) & 0\ \ \ \ \ \ \ \ \ \ \ \  (0.00\%) & 12592 & 12592 \\\hline
ContextualAssertion & absent, possible, associated with someone else & 3525 (100.00\%) & 0\ \ \ \ \ \ \ \ \ \ \ \  (0.00\%) & 0\ \ \ \ \ \ \ \ \ \ \ \  (0.00\%) & 3377 & 3377 \\\hline
GPT-4o & present, absent, possible, hypothetical, associated with someone else & 12444 (100.00\%) & 0 \ \ \ \ \ \ \ \ \ \ \ \ (0.00\%) & 0\ \ \ \ \ \ \ \ \ \ \ \ (0.00\%) & 12444 & 12444 \\\hline
Azure Ai Textanalytics & absent, possible, hypothetical, associated with someone else & 2714 (68.37\%) & 1150 (28.96\%) & 103\ \ \ \ \ \ \ \ \ \ \ \  (2.59\%) & 3867 & 3970 \\\hline
AWS Comprehend & present, absent, possible, hypothetical, associated with someone else & 7644 (61.41\%) & 4427 (35.57\%) & 373\ \ \ \ \ \ \ \ \ \ \ \  (3.00\%) & 12071 & 12444 \\\hline
Negex & absent & 1443 (55.64\%) & 535\ \ \ \  (20.63\%) & 616\ \ \ \ \ \ \ \ \ \ \ \  (23.73\%) & 1978 & 2594 \\
\hline
\end{tabular}
}
\begin{flushleft}
\footnotesize{\textit{*associated with someone else: Refers to medical conditions related to individuals other than the patient, such as family members.}}
\end{flushleft}
\label{Tab:mathing_rates}
\end{table}

\subsection*{A.3  Label Correspondence}
\begin{table}[htb!]
\scriptsize
\centering
\caption{Label Correspondences: AWS model does not represent \textit{conditional} labels. Azure AI Text Analytics for Health has 3 distinct \textit{possible} labels - \textit{positive}, \textit{negative} and \textit{neutral}, while \textit{present} is not represented. }
\label{tab:comprehensive_label_mapping}
\begin{tabular}{@{}lll@{}}
\toprule
\textbf{i2b2 Label} & \textbf{AWS Comprehend Label} & \textbf{Azure Ai Textanalytics} \\
\midrule
Present & SIGN/SYMPTOM/DIAGNOSIS & N/A \\
Absent & NEGATION & Negative \\
Possible & LOW\_CONFIDENCE &  Possible \\
Hypothetical & HYPOTHETICAL & Hypothetical \\
Conditional & N/A & Conditional \\
Associated\_with\_someone\_else & PERTAINS\_TO\_FAMILY & Other \\
\bottomrule
\end{tabular}
\end{table}

\subsection*{A.4 Latency Comparison}
\begin{table}[htb!] 
\scriptsize
\centering
\caption{Mean latency per 100 rows, measured in seconds for various assertion methods. Experiments were run on Google Colab servers, with CPU tasks performed on a CPU instance (8vCPU @ 2.2 GHz, 50.99 GB RAM) and GPU tasks executed on an NVIDIA A100 GPU (40 GB HBM2).}
\label{tab:speed_benchmarks}
\begin{tabular}{l c c c}
\hline
\textbf{Methods}            & \textbf{Parameter Size} & \textbf{CPU(seconds)} & \textbf{GPU(seconds)} \\
\hline
Fine-Tuned LLM              & 8 Billion              & N/A            & 294          \\
Combined Pipeline           & N/A                      & 12        &  4          \\
AssertionDL                 & 11 Million                    & 3        &  2         \\
FewShotAssertion            & 109 Million                     & 5       &  2          \\
ContextualAssertion         & N/A                      & 2        & 1          \\
BFSC                        & 110 Million                      &5           &4   \\ 
Negex                       & N/A                      & 1        & 1          \\
\hline
\end{tabular}
\end{table}

\subsection*{A.5 Pretrained Assertion Models in Healthcare NLP}
\begin{table}[htb!]
\centering
\tiny
\caption{Comparison of pretrained models across various assertion categories. Each row corresponds to a specific pretrained model—such as Clinical, Oncology, Radiology, Smoking, Menopause, and Social Determinants of Health (SDOH)—highlighting their strengths in different assertion labels. Overall, this comparison serves as a valuable resource for practitioners and researchers seeking to select the most appropriate model for their specific application in medical text analysis.
}
\label{tab:pretrained_models_comparison_transposed}
\resizebox{\textwidth}{!}{%
\begin{tabular}{l|c|c|c|c|c|c}
\hline
\textbf{Categories} & \textbf{Clinical} & \textbf{Oncology} & \textbf{Radiology} & \textbf{Smoking} & \textbf{Menapause} & \textbf{SDOH} \\
\hline
Present & 0.95 & 0.90 & - & 0.92 & 0.90 & 0.87 \\
Past & 0.91 & 0.93 & - & 0.91 & 0.43 & 0.71 \\
Possible & 0.86 & 0.77 & - & - & 0.52 & 0.74 \\
Absent & 0.97 & 0.86 & - & 0.97 & 0.84 & 0.94 \\
Hypothetical & 0.89 & 0.71 & - & - & 0.75 & 0.83 \\
Family & 0.92 & 0.92 & - & - & 0.57 & - \\
Someone Else & 0.94 & - & - & - & - & 0.81 \\
Planned & 0.82 & - & - & - & 0.73 & - \\
Conditional & 0.30 & - & - & - & - & - \\
Confirmed & - & - & 0.95 & - & - & - \\
Negative & - & - & 0.96 & - & - & - \\
Suspected & - & - & 0.86 & - & - & - \\
\hline
\end{tabular}%
}
\end{table}

\clearpage
\subsection*{A.6  GPT Prompt}
\label{subsection GPT Prompt}
\begin{figure}[h!]
\begin{tcolorbox}[colframe=blue!50!black, colback=blue!10!white, coltitle=black, title=GPT-4o Prompt]
\footnotesize
You are a highly experienced medical data expert specializing in patient medical records.\\

In this context, an assertion refers to the sentiment or condition associated with
a specific medical entity within the context of a patient's record. This helps
determine whether symptoms or conditions are present, absent, possible, hypothetical,
or related to someone else, enhancing the precision of medical documentation and analysis.\\

Your task is to detect the assertion status of medical conditions mentioned in notes.
The possible assertion types are:
\begin{itemize}
    \item **absent**: condition is explicitly negated
    \item **associated\_with\_someone\_else**: condition refers to someone other than the patient
    \item **conditional**: condition is mentioned as contingent on another factor
    \item **hypothetical**: condition is part of a hypothetical scenario
    \item **possible**: condition is suggested as a possibility but not confirmed
    \item **present**: condition is clearly present for the patient
\end{itemize}

\#\#\# Instructions:
\begin{itemize}
    \item[1] Analyze the input TEXT and identify the assertion status of the TARGET condition.
    \item[2] Format your answer in valid JSON, using double quotes for both keys and values.
    \item[3] If multiple assertions are required, choose the most confident one.
\end{itemize}

\#\#\# EXAMPLE INPUT

\{\\
  "TEXT": "She was then started on Heparin with transition to Coumadin (goal INR of 2-3 secondary to h/o bilateral DVTs).",\\
  "TARGET": "bilateral DVT"\\
\}\\

\#\#\# INPUT\\
\\
\{\\
  "TEXT": "{text}",\\
  "TARGET": "{target}"\\
\}\\

\#\#\# Your Answer in JSON:\\
Provide a JSON object where the text and assertion type are the key-value pairs.\\

Example Output Format:\\

\{\\
  "TARGET": "bilateral dvt",\\
  "ASSERTION\_STATUS": "present"\\
\}
\end{tcolorbox}
\caption{Example of GPT-4o prompt for detecting assertion status in medical records}
\label{fig:gpt_prompt}
\end{figure}

\clearpage
\subsection*{A.7 Fine-tuned LLM Prompt}
\label{subsection fine-tuned LLM Prompt}
\begin{figure}[h!]
\begin{tcolorbox}[colframe=blue!50!black, colback=blue!10!white, coltitle=black, title=Fine-tuned LLM Prompt]
\footnotesize
You are provided with a document and an extracted entity (chunk).\\

Your job is to analyze the document and the chunk, understand the context, and assign one of the following statuses to the chunk:\\

\begin{itemize}
    \item **present**: If the chunk is mentioned in the context of the person.  
          *Example*: "He has a fractured ankle."
    \item **absent**: If the chunk is explicitly negated by the person.  
          *Example*: "He did not suffer from pain." (In this case, "pain" is absent/negated.)
    \item **hypothetical**: If the chunk is mentioned in a hypothetical scenario or as part of guidelines.  
          *Example*: "Adults above 70 are at greater risk of cancer." (Here, "cancer" is hypothetical.)
    \item **possible**: If the chunk is mentioned in a way that implies possibility.  
          *Example*: "Possible fracture."
    \item **associated\_with\_someone\_else**: If the condition refers to someone other than the patient.  
          *Example*: "Her mother has breast cancer."
\end{itemize}

\#\#\# Document:\\
\{\\
  "DOCUMENT": "{doc}"\\
\}\\

\#\#\# Chunk:\\
\{\\
  "CHUNK": "{chunks}"\\
\}\\

\#\#\# Your Answer in JSON:\\
Provide a JSON object where the chunk and assertion status are the key-value pairs.\\

Example Output Format:\\

\{\\
  "CHUNK": "fractured ankle",\\
  "ASSERTION\_STATUS": "present"\\
\}
\end{tcolorbox}
\caption{Example of Fine-tuned LLM prompt for detecting assertion status in medical records}
\label{fig:finetuned_llm_prompt}
\end{figure}

\lstdefinestyle{mystyle}{
    backgroundcolor=\color{gray!10},  
    basicstyle=\ttfamily\footnotesize, 
    breaklines=true,  
    frame=single, 
    captionpos=b, 
    numbers=left, 
    numberstyle=\tiny, 
    keywordstyle=\color{blue}, 
    commentstyle=\color{gray}, 
    stringstyle=\color{red}, 
}

\newpage 
\subsection*{A.8 Healthcare NLP Pipeline}
\begin{lstlisting}[language=Python, style=mystyle]
from sparknlp.base import DocumentAssembler
from sparknlp.annotator import Tokenizer, WordEmbeddingsModel, AssertionDLModel
from sparknlp_jsl.annotator import AssertionChunkConverter

# Document Assembler
document_assembler = DocumentAssembler() \
    .setInputCol("text") \
    .setOutputCol("sentence")

# Tokenizer
tokenizer = Tokenizer() \
    .setInputCols(["sentence"]) \
    .setOutputCol("token")

# Assertion Chunk Converter
converter = AssertionChunkConverter() \
    .setInputCols("token")\
    .setChunkTextCol("chunk")\
    .setChunkBeginCol("begin")\
    .setChunkEndCol("end")\
    .setOutputTokenBeginCol("token_begin")\
    .setOutputTokenEndCol("token_end")\
    .setOutputCol("ner_chunk")

# Word Embeddings Model
word_embeddings_100 = WordEmbeddingsModel.pretrained("embeddings_healthcare_100d", "en", "clinical/models")\
    .setInputCols(["sentence", "token"])\
    .setOutputCol("embeddings")

# AssertionDL Model
clinical_assertion_100 = AssertionDLModel.pretrained("assertion_dl_healthcare", "en", "clinical/models") \
    .setInputCols(["sentence", "ner_chunk", "embeddings"]) \
    .setOutputCol("assertionDL")\
    .setEntityAssertionCaseSensitive(False)

#FewShotAssertion Model
few_shot_assertion_converter = FewShotAssertionSentenceConverter()\
    .setInputCols(["sentence", "token", "ner_chunk"])\
    .setOutputCol("assertion_sentence")

e5_embeddings = E5Embeddings.pretrained("e5_base_v2_embeddings_medical_assertion_i2b2", "en", "clinical/models")\
    .setInputCols(["assertion_sentence"])\
    .setOutputCol("assertion_embedding")

few_shot_assertion_classifier = FewShotAssertionClassifierModel()\
    .pretrained("fewhot_assertion_i2b2_e5_base_v2_i2b2", "en", "clinical/models")\
    .setInputCols(["assertion_embedding"])\
    .setOutputCol("assertion_fewshot")
    
\end{lstlisting}

\newpage 
\begin{lstlisting}[language=Python, style=mystyle]
#Contextual Assertion Models
contextual_assertion_possible = ContextualAssertion.pretrained("contextual_assertion_possible","en","clinical/models")\
    .setInputCols("sentence", "token", "ner_chunk") \
    .setOutputCol("ca_possible")

contextual_assertion_conditional = ContextualAssertion.pretrained("contextual_assertion_conditional","en","clinical/models")\
    .setInputCols("sentence", "token", "ner_chunk") \
    .setOutputCol("ca_conditional") 
#Merger
assertionMerger_fewshot = AssertionMerger()\
      .setInputCols("assertion_fewshot")\
      .setOutputCol("assertion_merger_fewshot")\
      .setWhiteList(["absent","hypothetical"])

assertionMerger_dl = AssertionMerger()\
      .setInputCols("assertionDL")\
      .setOutputCol("assertion_merger_dl")\
      .setWhiteList(["associated_with_someone_else","conditional"])

assertionMerger_all = AssertionMerger()\
      .setInputCols("assertionDL","assertion_fewshot","ca_possible")\
      .setOutputCol("assertion_merger_all")\
      .setMergeOverlapping(True)\
      .setMajorityVoting(False)\
      .setOrderingFeatures(["confidence"])\
      .setWhiteList(["present","possible"])\
      .setApplyFilterBeforeMerge(True)

assertionMerger_final = AssertionMerger()\
      .setInputCols("assertion_merger_fewshot","assertion_merger_dl","assertion_merger_all","ca_conditional")\
      .setOutputCol("assertion_merger")\
      .setMergeOverlapping(True)\
      .setMajorityVoting(True)\
      .setOrderingFeatures(["confidence"])\
#Pipeline
pipeline = Pipeline(stages=[
            document_assembler,
            tokenizer,
            converter,
            few_shot_assertion_converter,
            e5_embeddings,
            few_shot_assertion_classifier,
            word_embeddings_100,
            clinical_assertion_100,
            assertionMerger_fewshot,
            contextual_assertion_conditional,
            contextual_assertion_possible,
            assertionMerger_dl,
            assertionMerger_all,
            assertionMerger_final
        ])
empty_data = spark.createDataFrame([[""]]).toDF("text")
dataframe = spark.createDataFrame(data_df)
model = pipeline.fit(empty_data)
result = model.transform(dataframe)    
\end{lstlisting}

\end{document}